\documentclass[11pt,a4paper]{article}
\usepackage[hyperref]{acl2020}

\usepackage{times}
\usepackage{latexsym}

\usepackage{microtype}

\aclfinalcopy 

\usepackage{tabularx}
\usepackage{booktabs}
\usepackage{makecell}
\usepackage{graphicx}

\title{Probing Neural Dialog Models for Conversational Understanding}

\author{Abdelrhman Saleh$^{1}$, 
 Tovly Deutsch$^{1,}$\thanks{\hspace{1.5 mm}Second author equal contribution.}, 
 Stephen Casper$^{1,}\footnotemark[1]$,
\\\textbf{Yonatan Belinkov$^{1,2}$,} and \textbf{Stuart Shieber$^{1}$}\\
$^1$Harvard School of Engineering and Applied Sciences\\
$^2$MIT Computer Science and Artificial Intelligence Laboratory\\
\texttt{abdelrhman\_saleh@college.harvard.edu}}

\date{}

\begin{document}
\maketitle
\begin{abstract}
The predominant approach to open-domain dialog generation relies on end-to-end training of neural models on chat datasets. However, this approach provides little insight as to what these models learn (or do not learn) about engaging in dialog. In this study, we analyze the internal representations learned by neural open-domain dialog systems and evaluate the quality of these representations for learning basic conversational skills. Our results suggest that standard open-domain dialog systems struggle with answering questions, inferring contradiction, and determining the topic of conversation, among other tasks. We also find that the dyadic, turn-taking nature of dialog is not fully leveraged by these models. By exploring these limitations, we highlight the need for additional research into architectures and training methods that can better capture high-level information about dialog.\footnote{Our code is available at \url{https://github.com/AbdulSaleh/dialog-probing}}
\end{abstract}

\section{Introduction}
Open-domain dialog systems often rely on neural models for language generation that are trained end-to-end on chat datasets. End-to-end training eliminates the need for hand-crafted features and task-specific modules (for example, for question answering or intent detection), while delivering promising results on a variety of language generation tasks including machine translation \cite{bahdanau2014neural}, abstractive summarization \cite{rush2015neural}, and text simplification \cite{wang2016simplification}. 

However, current generative models for dialog suffer from several shortcomings that limit their usefulness in the real world. Neural models can be opaque and difficult to interpret, posing barriers to their deployment in safety-critical applications such as mental health or customer service \cite{belinkov2019analysis}. End-to-end training provides little insight as to what these models learn about engaging in dialog. Open-domain dialog systems also struggle to maintain basic conversations, frequently ignoring  user input \cite{sankar2019neural} while generating irrelevant, repetitive, and contradictory responses \cite{saleh2019hierarchical,li2016deep,li2017adversarial,welleck2018dnli}. Table \ref{tab:badsamples} shows examples from standard dialog models which fail at basic interactions -- struggling to answer questions, detect intent, and understand conversational context. 

In light of these limitations, we aim to answer the following questions: (\textit{i}) Do neural dialog models effectively encode information about the conversation history? (\textit{ii}) Do neural dialog models learn basic conversational skills through end-to-end training? (\textit{iii}) And to what extent do neural dialog models leverage the dyadic, turn-taking structure of dialog to learn these skills? 

To answer these questions, we propose a set of eight \emph{probing tasks} to measure the conversational understanding of neural dialog models. Our tasks include question classification, intent detection, natural language inference, and commonsense reasoning, which all require high-level understanding of language. We also carry out \emph{perturbation experiments} designed to test if these models fully exploit dialog structure during training. These experiments entail breaking the dialog structure by training on shuffled conversations and measuring the effects on probing performance and perplexity. 

We experiment with both recurrent \cite{sutskever2014sequence} and transformer-based \cite{vaswani2017attention} open-domain dialog models. We also analyze models with different sizes and initialization strategies, training small models from scratch and fine-tuning large pre-trained models on dialog data. Thus, our study covers a variety of standard models and approaches for open-domain dialog generation. Our analysis reveals three main insights:
\begin{enumerate}
    \item Dialog models trained from scratch on chat datasets perform poorly on the probing tasks, suggesting that they struggle with basic conversational skills. Large, pre-trained models achieve much better probing performance but are still on par with simple baselines. 

    \item Neural dialog models fail to effectively encode information about the conversation history and the current utterance. In most cases, simply averaging the word embeddings is superior to using the learned encoder representations. This performance gap is smaller for large, pre-trained models.
    
    \item Neural dialog models do not leverage the dyadic, turn-taking nature of conversation. Shuffling conversations in the training data had little impact on perplexity and probing performance. This suggests that breaking the dialog structure did not significantly affect the quality of learned representations. 
\end{enumerate}

Our code integrates with and extends ParlAI \cite{miller2017parlai}, a popular open-source platform for building dialog systems. We also publicly release all our code at \url{https://github.com/AbdulSaleh/dialog-probing}, hoping that probing will become a standard method for interpreting and analyzing open-domain dialog systems.

\begin{table*}[th]
\centering
\begin{tabular}{p{3cm}|p{3.3cm}|p{3.4cm}|p{3.05cm}}
\normalsize 
\makecell[lp{2.6cm}]{Question \\Answering} & \makecell[lp{2.6cm}]{Inference \& \\Entailment} &  \makecell[lp{3.0cm}]{Intent \\Detection} & \makecell[lp{2.25cm}]{Emotional \\Understanding} \\
\midrule

\makecell[lp{3.0cm}]{[Usr]: What do you do for a living?
\\ {[}Bot{]}: I'm from south america, lol
} & 

\makecell[lp{3.3cm}]{[Usr]: what's your \\favorite food?
\\ {[}Bot{]}: I love food
\\{[}Usr{]}: you like food?
\\{[}Bot{]}: no, I don't }& 

\makecell[lp{3.4cm}]{[Usr]: Give me music recommendations.
\\ {[}Bot{]}: I don't know what I want to do with my life.
} & 

\makecell[lp{3.1cm}]{[Usr]: I feel so\\ miserable
\\ {[}Bot{]}: I'm afraid it's too much for me.} \\

\bottomrule
\end{tabular}
\caption{Examples from open-domain dialog systems struggling with simple conversational skills.}
\label{tab:badsamples}
\end{table*}

\section{Related Work}

Evaluating and interpreting open-domain dialog models is notoriously challenging. Multiple studies have shown that standard evaluation metrics such as perplexity and BLEU scores \cite{papineni2002bleu}
correlate very weakly with human judgements of conversation quality \cite{liu2016not,ghandeharioun2019approximating,dziri2019evaluating}. This has inspired multiple new approaches for evaluating dialog systems. One popular evaluation metric involves calculating the semantic similarity between the user input and generated response in high-dimensional embedding space \cite{liu2016not,ghandeharioun2019approximating,dziri2019evaluating,park2018hierarchical,zhao2017learning,xu2018towards}. \citet{ghandeharioun2019approximating} proposed calculating conversation metrics such as sentiment and coherence on self-play conversations generated by trained models. Similarly, \citet{dziri2019evaluating} use neural classifiers to identify whether the model-generated responses entail or contradict user input in a natural language inference setting. 

To the best of our knowledge, all existing approaches for evaluating the performance of open-domain dialog systems only consider external model behavior in the sense that they analyze properties of the generated text. In this study, we explore internal representations instead, motivated by the fact that reasonable internal behavior is crucial for interpretability and is often a prerequisite for effective external behavior.

Outside of open-domain dialog, probing has been applied for analyzing natural language processing models in machine translation \cite{belinkov2017neural} and visual question answering \cite{subramaniananalyzing}. Probing is also commonly used for evaluating the quality of ``universal'' sentence representations which are trained once and used for a variety of applications \cite{conneau2018you,adi2016fine} (for example, InferSent \cite{conneau2017supervised}, SkipThought \cite{kiros2015skip}, USE \cite{cer2018universal}). Along the same lines, natural language understanding benchmarks such as GLUE \cite{wang2018glue} and SuperGLUE \cite{wang2019superglue} propose a set of diverse tasks for evaluating general linguistic knowledge. Our analysis differs from previous work since it is focused on probing for conversational skills that are particularly relevant to dialog generation.

With regard to perturbation experiments, \citet{sankar2019neural} found that standard dialog models are largely insensitive to perturbations of the input text. Here we introduce an alternative set of perturbation experiments to similarly explore the extent to which dialog structure is being leveraged by these models.

\section{Methodology}
\subsection{Models and Data}

In this study, we focus on the three most widespread dialog architectures: recurrent neural networks (RNNs) \cite{sutskever2014sequence}, RNNs with attention \cite{bahdanau2014neural}, and Transformers \cite{vaswani2017attention}. We use the ParlAI platform \cite{miller2017parlai} for building and training the models. We train models of two different sizes and initialization strategies. Small models ($\approx 14$M parameters) are initialized randomly and trained from scratch on DailyDialog \cite{dailydialog}. Large models ($\approx70$M parameters) are pre-trained on WikiText-103 \cite{merity2016wikitext}, and then fine-tuned on DailyDialog.\footnote{See the supplemental material for further training details.} 

DailyDialog \cite{dailydialog} is a dataset of $14$K train, $1$K validation, and $1$K test multi-turn dialogs collected from an English learning website. The dialogs are of much higher quality than datasets scraped from Twitter or Reddit. WikiText-103 \cite{merity2016wikitext} is a dataset of $29$K Wikipedia articles. For pre-training the large models, we format WikiText-103 as a dialog dataset by treating each paragraph as a conversation and each sentence as an utterance.

\subsection{Probing experiments} \label{sec:probingexps}
\begin{figure}
\centering
\includegraphics[width=\linewidth]{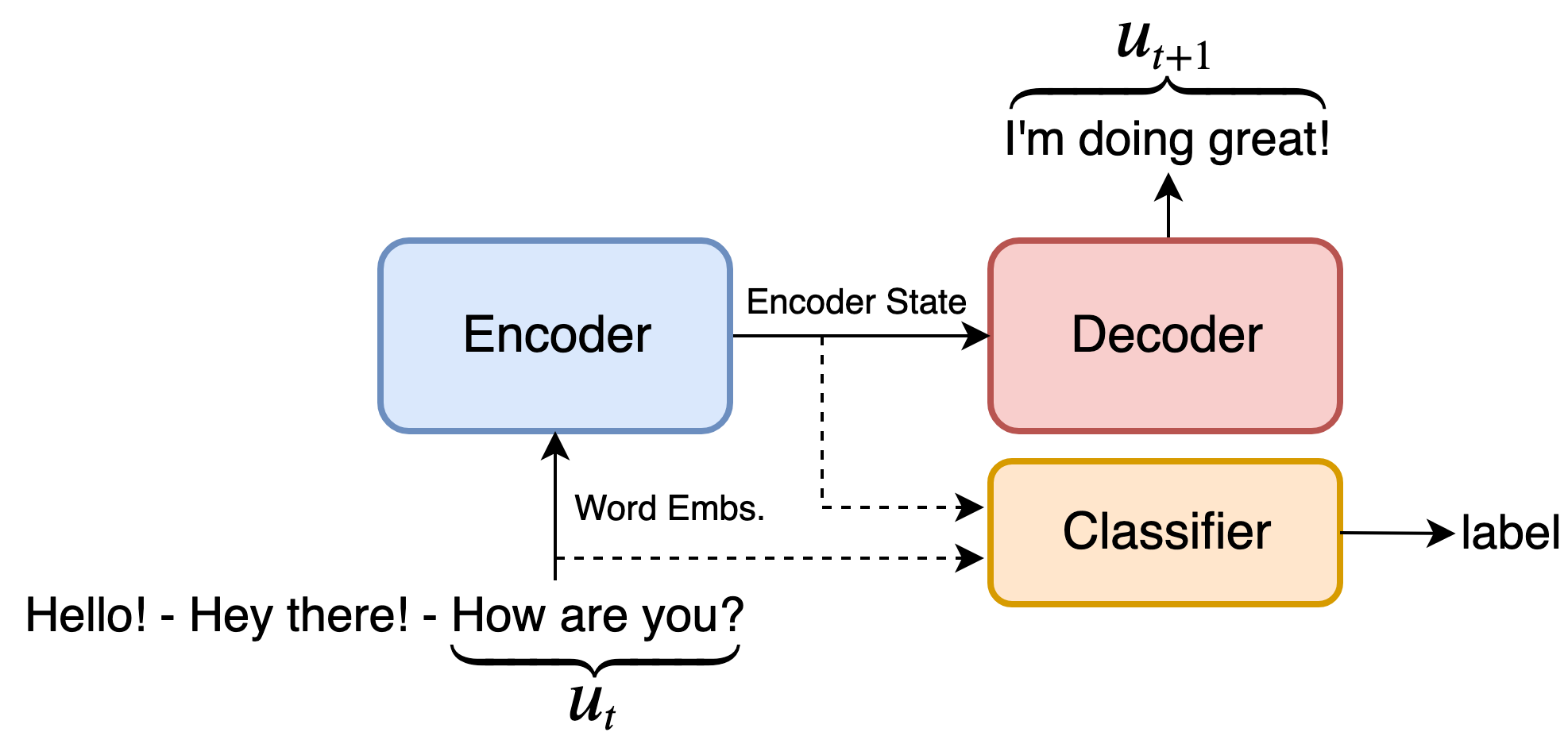}
\caption{Probing setup. Dotted arrows emphasize that probing is applied to frozen models after dialog training. Only the parameters of the classifier module are learned during probing.}
\label{fig:probing}
\end{figure}

In open-domain dialog generation, the goal is to generate the next utterance or response, $u_{t+1}$, given the conversation history, $[u_1, \dots, u_t]$. First, we train our models on dialog generation using a maximum-likelihood objective \cite{sutskever2014sequence}. We then freeze these trained models and use them as feature extractors. We run the dialog models on text from the probing tasks and use the internal representations as features for a two-layer multilayer perceptron (MLP) classifier trained on the probing tasks as in figure \ref{fig:probing}. This follows the same methodology outlined in previous probing studies \cite{belinkov2017neural,belinkov2017speech,conneau2018you,adi2016fine}.

The assumption here is that if a model learns certain conversational skills, then knowledge of these skills should be reflected in its internal representations. For example, a model that excels at answering questions would be expected to learn useful internal representations for question answering. Thus, the performance of the probing classifier on question answering can be used as a proxy for learning this skill. We extend this reasoning to eight probing tasks designed to measure a model's conversational understanding.

The probing tasks require high-level reasoning, sometimes across multiple utterances, therefore we aggregate utterance-level representations for probing. Our probing experiments consider three types of internal representations: 

\paragraph{Word Embeddings:} To get the word embedding representations, we first averaged word embeddings of all words in the previous utterances, $[u_1, \dots, u_{t-1}]$, then we separately averaged word embeddings of all words in the current utterance, $u_t$, and concatenated the two resulting, equal-length vectors. Encoding the past utterances and the current utterance separately is important since it provides some temporal information about utterance order. We used the dialog model's encoder word embedding matrix.

\paragraph{Encoder State:}  For the the encoder state, we extracted the encoder outputs after running it on the entire probing task input (i.e. the full conversation history, $[u_1, \dots, u_t]$). Crucially, encoder states are the representations passed to the decoder for generation and are thus different for each architecture. For RNNs we used the \textit{last} encoder hidden and cell states. For RNNs with attention the decoder has access to all the encoder hidden states (not just the final ones), through the attention mechanism. Thus, for RNNs with attention, we first averaged the encoder hidden states corresponding to the previous utterances, $[u_1, \dots, u_{t-1}]$, and then we separately averaged the encoder hidden states corresponding to the current utterance, $u_t$, and concatenated the two resulting, equal-length vectors. We also concatenated the last cell state. Similarly, for Transformers, we averaged the encoder outputs corresponding to the previous utterances and separately averaged encoder outputs corresponding to the current utterance and concatenated them. 

\paragraph{Combined:} The combined representations are the concatenation of of the word embeddings and encoder state representations.

\paragraph{}We also use GloVe \cite{pennington2014glove} word embeddings as a simple baseline. We encode the probing task inputs using the word embeddings approach described above. We ensure that GloVe and all models of a certain size (small vs large) share the same vocabulary for comparability.

\subsection{Perturbation Experiments}
We also propose a set of perturbation experiments designed to measure whether dialog models fully leverage dialog structure for learning conversational skills. We create a new training dataset by shuffling the order of utterances within each conversation in DailyDialog. This completely breaks the dialog structure and utterances no longer naturally follow one another. We train (or fine-tune) separate models on the shuffled dataset and evaluate their probing performance relative to models trained on data as originally ordered.

\section{Probing Tasks} \label{sec:probingtasks}

The probing tasks selected for this study measure conversational understanding and skills relevant to dialog generation. Some tasks are inspired by previous benchmarks \cite{wang2018glue}, while others have not been explored before for probing. Examples are listed in the supplemental material.

\paragraph{TREC:} Question answering is a key skill for effective dialog systems. A system that deflects user questions could seem inattentive or indifferent. In order to correctly respond to questions, a model needs to determine what type of information the question is requesting. We probe for question answering using the TREC question classification dataset \cite{li2002learning}, which consists of questions labeled with their associated answer types.

\paragraph{DialogueNLI:} Any two turns in a conversation could entail each other (speakers agreeing, for example), or contradict each other (speakers disagreeing), or be unrelated (speakers changing topic of conversation). A dialog system should be sensitive to contradictions to avoid miscommunication and stay aligned with human preferences. We use the Dialogue NLI dataset \cite{welleck2018dnli}, which consists of pairs of dialog turns with entailment, contradiction, and neutral labels to probe for natural language inference. The original dataset examines two utterances from the same speaker (``I go to college", ``I am a student"), so we modify the second utterance to simulate a second speaker (``I go to college", ``You are a student").

\paragraph{MultiWOZ:} Every utterance in a conversation can be considered as an action or a dialog act performed by the speaker. A speaker could be making a request, providing information, or simply greeting the system. MultiWOZ 2.1 \cite{eric2019multiwoz} is a dataset of multi-domain, goal-oriented conversations. Human turns are labeled with dialog acts and the associated domains (hotel, restaurant, etc.), which we use to probe for natural language understanding. 

\paragraph{SGD:} Tracking user intent is also important for generating appropriate responses. The same intent is often active across multiple dialog turns since it takes more than one turn to book a hotel, for example. Determining user intent requires reasoning over multiple turns in contrast to dialog acts which are turn-specific. To probe for this task, we use intent labels from the multi-domain, goal-oriented Schema-Guided Dialog dataset \cite{rastogi2019schema}. 

\paragraph{WNLI:} Endowing neural models with commonsense reasoning is an ongoing challenge in machine learning \cite{storks2019commonsense}. We use the Winograd NLI dataset, a variant of the Winograd Schema Challenge \cite{levesqueWinogradSchemaChallenge2012}, provided in the GLUE benchmark \cite{wang2018glue} to probe for commonsense reasoning. WNLI is a sentence pair classification task where the goal is to identify whether the hypothesis correctly resolves the referent of an ambiguous pronoun in the premise. 

\paragraph{SNIPS:} The Snips NLU benchmark \cite{coucke2018snips} is a dataset of crowdsourced, single-turn queries labeled for intent. We use this dataset to probe for intent classification. 

\paragraph{ScenarioSA:} An understanding of sentiment and emotions is crucial for building social, human-centered conversational agents. We use ScenarioSA \cite{zhang2019scenariosa} as a sentiment classification probing task. The dataset is composed of natural, multi-turn, open-ended dialogs with turn-level sentiment labels. 

\paragraph{DailyDialog Topic:} The DailyDialog dataset comes with conversation-level annotations for ten diverse topics, such as ordinary life, school life, relationships, and health. Inferring the topic of conversation is an important skill that could help dialog systems stay consistent and on topic. We use dialogs from the DailyDialog test set to create a probing tasks where the goal is to classify a dialog into the appropriate topic.

\section{Results}

\begin{figure}[h]
    \centering
    \includegraphics[width=\linewidth]{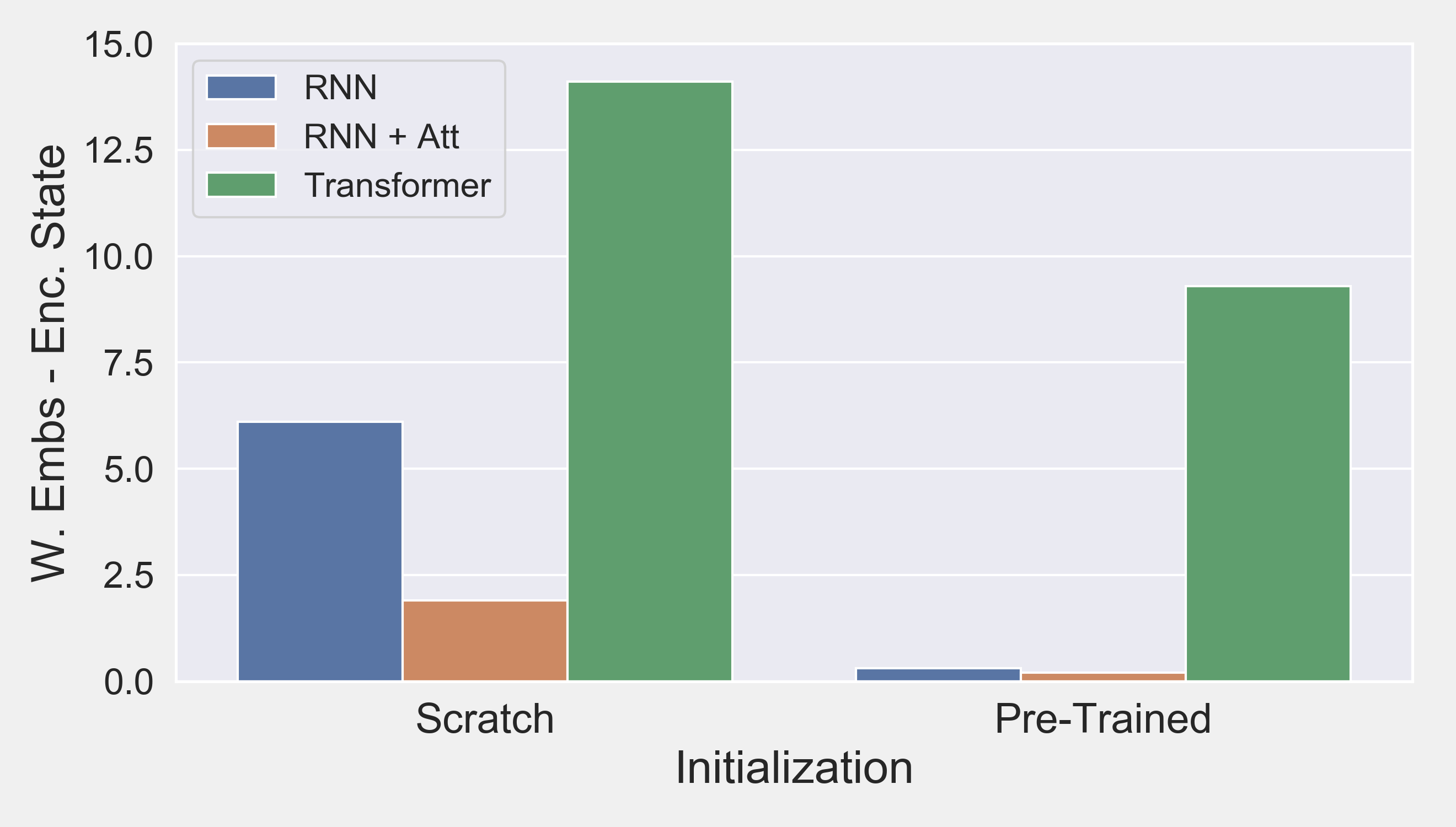}
    \caption{Bar plot showing difference between average scores for word embeddings and encoder states.}
    \label{fig:wembs_vs_context}
\end{figure}

\subsection{Quality of Encoder Representations} \label{sec:qualityrepresentations}
Results from our probing experiments are presented in tables \ref{tab:dailydialogprobes} and \ref{tab:wikiprobes}. We calculate an average score to summarize the overall accuracy on all tasks. Here we explore whether the encoder learns high quality representations of the conversation history. We focus on \textit{encoder states} because these representations are passed to the decoder and used for generation (figure \ref{fig:probing}). Thus, effectively encoding information in the encoder states is crucial for dialog generation. 

Figure \ref{fig:wembs_vs_context} shows the difference in average probing accuracy between the word embeddings and the encoder state for each model. The word embeddings outperform the encoder state for all the small models. This performance gap is most pronounced for the Transformer but is non-existent for the large recurrent models.

One possible explanation is that the encoder highlights information relevant to generating dialog at the cost of obfuscating or losing information relevant to the probing tasks -- given that the goals of certain probing tasks do not perfectly align with natural dialog generation. For example, the DailyDialog dataset contains examples where a question is answered with another question (perhaps for clarification). The TREC question classification task does not account for such cases and expects each question to have a specific answer type. This explanation is supported by the observation that the information in the word embeddings and encoder state is not necessarily redundant. The combined representations often outperform using either one separately (albeit by a minute amount). 

Regardless of the reason behind this gap in performance, multiple models still fail to effectively encode information about the conversation history that is already present in the word embeddings.

\begin{table*}
\centering
\noindent\makebox[\textwidth]{
\begin{tabular}{l|cccccccc|c}\toprule
\multicolumn{1}{@{} l|}{Model} & TREC & DNLI & MWOZ & SGD & SNIPS & WNLI & SSA & Topic & Avg \\ \midrule

\multicolumn{1}{@{} l|}{\textbf{Majority}}& 18.8	& 34.5 & 17.0 & 6.5 & 14.3 & 56.3 & 37.8 & 34.7 & 27.5  \\
\multicolumn{1}{@{} l|}{\textbf{GloVe Mini}}& 83.8 & 70.8 & 91.9 & 71.2 & 98.0 & 48.2 & 75.3& 54.0 & 74.2   \\

\hline \hline
\multicolumn{1}{@{} l}{\textbf{RNN}}  \\
Word Embs. &79.0 & 63.7 & 88.1 & 63.2 & 95.7 & 52.2 & 66.7 & 55.4 & \underline{65.7}\\
Enc. State  &80.4 & 55.4 & 69.7 & 47.3 & 93.4 & 49.4 & 62.5 & 56.8 & 60.2\\
Combined &81.9 & 60.0 & 82.4 & 60.9 & 95.3 & 49.9 & 64.8 & 57.3  & 64.4\\

\multicolumn{1}{@{} l}{\textbf{RNN + Attn}} \\
Word Embs. &75.6 & 64.5 & 87.5 & 65.9 & 96.5 & 50.1 & 62.6 & 55.1 & 69.7\\
Enc. State  &77.2&	59.5&	80.0	&57.0&	95.1&	49.9&	64.7&	59.0 & 67.8\\
Combined &79.2&	64.6&	86.3&	66.8&	96.7&	51.3&	65.3&	58.5 & \underline{71.1}\\

\multicolumn{1}{@{} l }{\textbf{Transformer}}    \\
Word Embs. &81.2&	71.6&	90.9&	70.9&	97.7	&48.6&	74.4&	62.3&	\textbf{\underline{74.7}}\\
Enc. State  &67.9&	54.1&	68.7&	47.2&	85.1&	49.4&	57.4&	55.4&	60.7\\
Combined   &81.5&	71.3&	91.2&	70.3&	97.9&	50.1&	72.8&	59.6&	74.3\\
\bottomrule
\end{tabular}}
\caption{Accuracy on probing tasks for small models trained with random initialization on DailyDialog. Best Avg result for each model underlined. Best Avg result in bold.}
\label{tab:dailydialogprobes}
\end{table*}

\begin{table*}
\centering
\noindent\makebox[\textwidth]{
\begin{tabular}{l|cccccccc|c}\toprule
\multicolumn{1}{@{} l |}{Model} & TREC & DNLI & MWOZ & SGD & SNIPS & WNLI & SSA & Topic & Avg\\ \midrule

\multicolumn{1}{@{} l|}{\textbf{Majority}}& 18.8	& 34.5 & 17.0 & 6.5 & 14.3 & 56.3 & 37.8 & 34.7 & 27.5  \\
\multicolumn{1}{@{} l|}{\textbf{GloVe}}& 86.5	&70.3	&91.6	&70.5	&97.8	&49.9	&75.1	&54.3 & 74.5  \\

\hline \hline
\multicolumn{1}{@{} l}{\textbf{RNN}}  \\
Word Embs.  &84.0 & 71.6 & 91.4 & 69.8 & 98.1 & 51.4 & 72.0 & 52.3 & 73.8\\
Enc. State  &84.6 & 66.8 & 89.9 & 72.9 & 97.2 & 48.6 & 67.8 & 61.0 & 73.6\\
Combined &85.6 & 69.4 & 91.1 & 74.0 & 97.6 & 49.6 & 69.1 & 61.4 & \underline{74.7}\\

\multicolumn{1}{@{} l}{\textbf{RNN + Attn}} \\
Word Embs.  &83.4 & 71.4 & 91.8 & 70.1 & 97.9 & 49.5 & 72.1 & 55.7 & 74.0\\
Enc. State   &  85.0&	65.6&	90.0&	73.6&	97.2&	47.5&	70.4&	63.0&	74.0 \\
Combined  & 86.6&	70.0&	92.0	&75.9&	97.7&	48.8&	73.5&	62.3&	\underline{75.9}  \\

\multicolumn{1}{@{} l}{\textbf{Transformer}}    \\
Word Embs.  &89.4&	70.4&	91.4&	70.3&	98.3&	51.4&	71.7&	51.5&	74.3\\
Enc. State  &71.3&	58.5&	70.7&	57.5&	88.5&	50.2&	58.8&	64.1&	65.0 \\
Combined & 90.0&	70.2&	91.1&	70.5&	98.1&	50.4&	72.4&	62.9&	\textbf{\underline{75.7}}\\

\bottomrule
\end{tabular}}
\caption{Accuracy on probing tasks for large, Wikipedia pre-trained models fine-tuned on DailyDialog. Best Avg result for each model underlined. Best Avg result in bold.}
\label{tab:wikiprobes}
\renewcommand{\arraystretch}{1.0}
\end{table*}

\begin{figure}
    \centering
    \includegraphics[width=\linewidth]{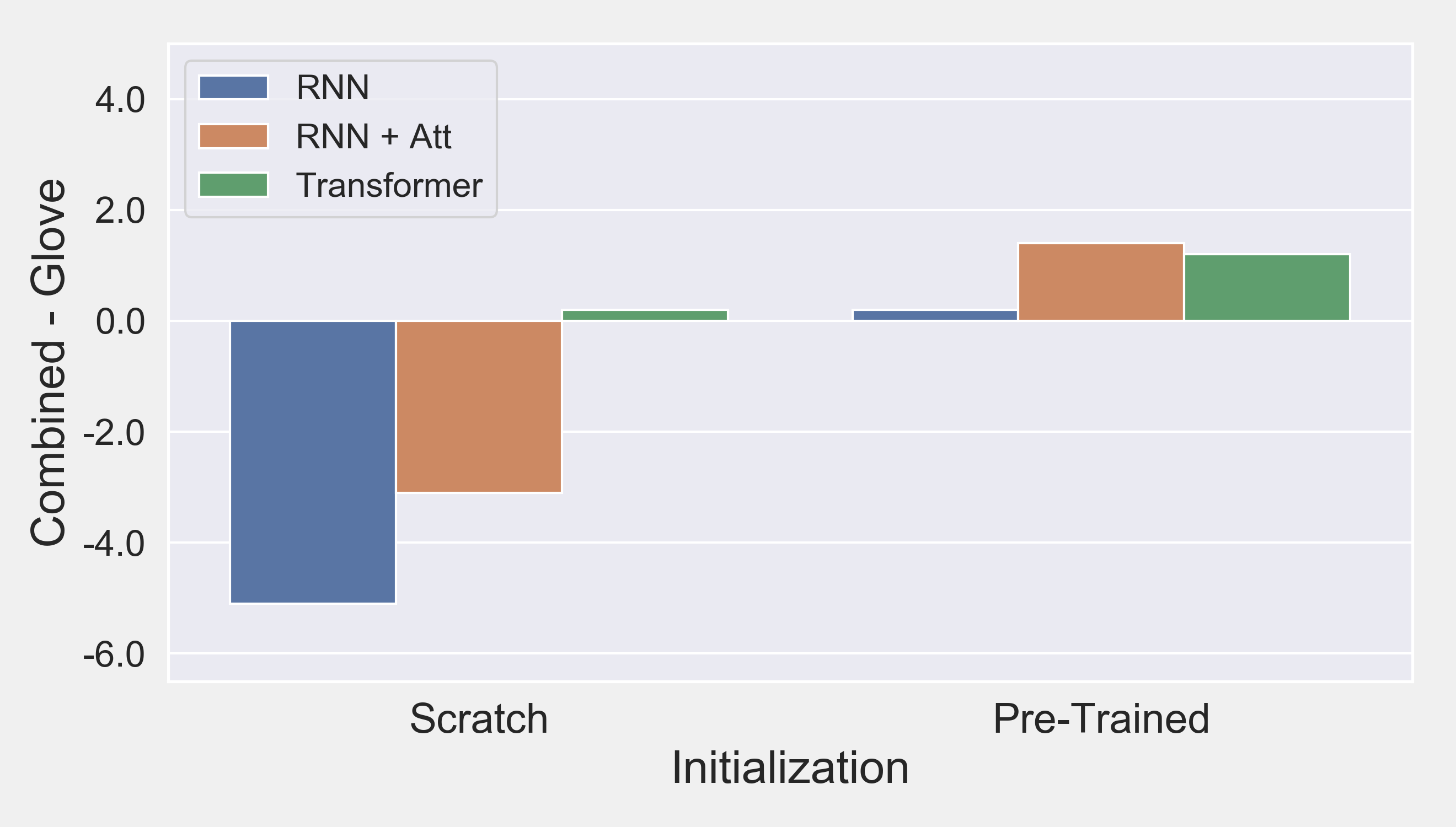}
    \caption{Bar plot showing difference between average scores for combined representations (word embeddings + encoder state) and GloVe baseline.}
    \label{fig:combined_vs_glove}
\end{figure}

\subsection{Probing for Conversational Understanding}
In this section, we compare the probing performance of the ordered dialog models to the simple baseline of averaging GloVe word embeddings. Here we consider the \textit{combined representations} since they achieve the best performance overall and can act as a proxy for all the information captured by the encoder about the conversation history.

Since our probing tasks test for conversational skills important for dialog generation, we would expect the dialog models to outperform GloVe word embeddings. However, this is generally not the case. As figure \ref{fig:combined_vs_glove} shows, the GloVe baseline outperforms the small recurrent models while being on par with the large pre-trained models in terms of average score. Tables \ref{tab:dailydialogprobes} and \ref{tab:wikiprobes} show that this pattern also generally applies at the task level, not just in terms of average score.

Closer inspection, however, reveals one exception. Combined representations from both the small and large models consistently outperform GloVe on the DailyDialog Topic task. This is the only task that is derived from the DailyDialog test data, which follows the same distribution as the dialogs used for training the models. This suggests that lack of generalization can partly explain the weak performance on other tasks. It is also worth noting that DailyDialog Topic is labeled at the conversation level rather than the turn level. Thus, identifying the correct label does not necessarily require reasoning about turn-level interactions (unlike DialogueNLI, for example). 

The poor performance on the majority of tasks, relative to the simple GloVe baseline, leads us to conclude that standard dialog models trained from scratch struggle to learn the basic conversational skills examined here. Large, pre-trained models do not seem to master these skills either, with performance on par with the baselines.

\begin{table*}
\centering
\noindent\makebox[\textwidth]{
\begin{tabular}{l|c|cccccccc|c}\toprule
\multicolumn{1}{@{} l|}{Model}& Test PPL & TREC & DNLI & MWOZ & SGD & SNIPS & WNLI & SSA & Topic & Avg \\ \midrule

\multicolumn{1}{@{} l|}{\textbf{Majority}}& - & 18.8	& 34.5 & 17.0 & 6.5 & 14.3 & 56.3 & 37.8 & 34.7 & 27.5   \\
\multicolumn{1}{@{} l|}{\textbf{GloVe Mini}}& - & 83.8 & 70.8 & 91.9 & 71.2 & 98.0 & 48.2 & 75.3& 54.0 & 74.2   \\

\hline \hline
\multicolumn{1}{@{} l}{\textbf{RNN}}  \\
Ordered & 27.2 & 80.4 & 55.4 & 69.7 & 47.3 & 93.4 & 49.4 & 62.5 & 56.8 & \underline{60.2} \\
Shuffled & 29.0 & 77.3 & 55.7 & 71.2 & 46.4 & 92.8 & 51.5 & 57.0 & 56.8 & 59.7\\

\multicolumn{1}{@{} l}{\textbf{RNN + Attn}} \\
Ordered & 26.0 &77.2&	59.5&	80.0	&57.0&	95.1&	49.9&	64.7&	59.0 & 67.8\\
Shuffled & 28.8 &80.2&	60.8&	80.8&	60.7&	92.9&	50.8&	57.9&	59.3&	\underline{67.9}\\

\multicolumn{1}{@{} l}{\textbf{Transformer}}    \\
Ordered & 29.3 &67.9&	54.1&	68.7&	47.2&	85.1&	49.4&	57.4&	55.4&	\underline{60.7}\\
Shuffled & 30.8  &58.6&	52.1	&62.6&	46.4&	83.5&	50.4&	53.5&	63.8&	58.9\\

\bottomrule
\end{tabular}}
\caption{Perplexity and accuracy on probing tasks for small models trained with random initialization on ordered and shuffled dialogs from DailyDialog. Results shown are for probing the encoder state. Best Avg result for each model underlined.}
\label{tab:dailydialogshuffles}
\vspace{-1.5cm}
\renewcommand{\arraystretch}{1.0}
\end{table*}

\begin{table*}
\centering
\noindent\makebox[\textwidth]{
\begin{tabular}{l|c|cccccccc|c}\toprule
\multicolumn{1}{@{} l|}{Model}& Test PPL & TREC & DNLI & MWOZ & SGD & SNIPS & WNLI & SSA & Topic & Avg \\ \midrule

\multicolumn{1}{@{} l|}{\textbf{Majority}}& - & 18.8	& 34.5 & 17.0 & 6.5 & 14.3 & 56.3 & 37.8 & 34.7 & 27.5   \\
\multicolumn{1}{@{} l|}{\textbf{GloVe}}& - & 86.5	&70.3	&91.6	&70.5	&97.8	&49.9	&75.1	&54.3  & 74.5 \\

\hline \hline
\multicolumn{1}{@{} l}{\textbf{RNN}}  \\
Ordered & 17.0 &84.6 & 66.8 & 89.9 & 72.9 & 97.2 & 48.6 & 67.8 & 61.0  & \underline{73.6}\\
Shuffled & 19.1  &85.4 & 65.1 & 89.5 & 69.0 & 97.3 & 50.5 & 64.7 & 65.4 & 73.4\\

\multicolumn{1}{@{} l}{\textbf{RNN + Attn}} \\
Ordered & 16.5 &85.0&	65.6&	90.0&	73.6&	97.2&	47.5&	70.4&	63.0&	\underline{74.0} \\
Shuffled & 19.6 &84.1&	64.9&	89.9&	71.1&	96.6&	50.3&	64.7&	65.4&	73.4\\

\multicolumn{1}{@{} l}{\textbf{Transformer}}    \\
Ordered & 19.8 &71.3&	58.5&	70.7&	57.5&	88.5&	50.2&	58.8&	64.1&	\underline{65.0}\\
Shuffled &21.4 &66.1&	58.0&	68.8&	58.0&	89.6&	49.0&	56.3&	64.2&	63.8
\\

\bottomrule
\end{tabular}}
\caption{Perplexity and accuracy on probing tasks for large, Wikipedia pre-trained models fine-tuned on ordered and shuffled dialogs from DailyDialog. Results shown are for probing the encoder state. Best Avg result for each model underlined.}
\label{tab:wikishuffles}
\vspace{-2.0cm}
\renewcommand{\arraystretch}{1.0}
\end{table*}

\begin{figure}
    \centering
    \includegraphics[width=\linewidth]{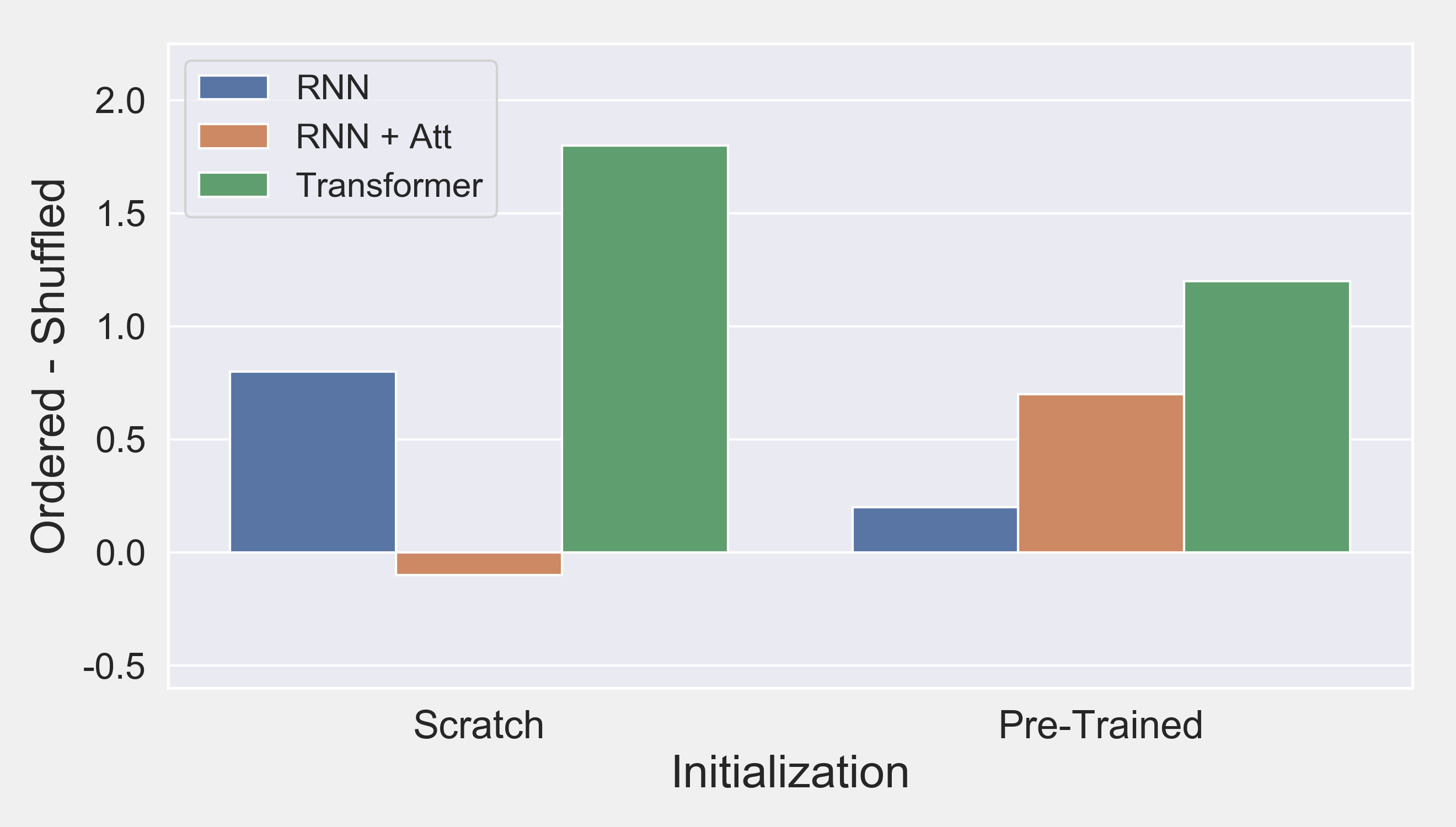}
    \caption{Bar plot showing difference between average scores for models trained on ordered and shuffled data.}
    \label{fig:ordered_vs_shuffled}
\end{figure}

\subsection{Effect of Dialog Structure}
Tables \ref{tab:dailydialogshuffles} and \ref{tab:wikishuffles} summarize the results of the perturbation experiments. Figure \ref{fig:ordered_vs_shuffled} shows the difference in average performance between the ordered and shuffled models. We show results for the \textit{encoder states} since these representations are important for encoding the conversation history, as discussed in section \ref{sec:qualityrepresentations}. The encoder states are also sensitive to word and utterance order, unlike averaging the word embeddings. So if a model can fully exploit the dyadic, turn-taking, structure of dialog, this is likely to be reflected in the encoder state representations.

In most of our experiments, models trained on ordered data outperformed models trained on shuffled data, as expected. We can see in figure \ref{fig:ordered_vs_shuffled}, that average scores for ordered models were often higher than for shuffled models. However, the absolute gap in performance was at most $2\%$, which is a minute difference in practice. And even though ordered models achieved higher accuracy on average, if we examine individual tasks in tables \ref{tab:dailydialogshuffles} and \ref{tab:wikishuffles}, we can find instances where the shuffled models outperformed the ordered ones for each of the tested architectures, sizes, and initialization strategies. 

The average difference in test perplexity between all the ordered and shuffled models was less than $2$ points. This is also a minor difference in practice, suggesting that model fit and predictions are not substantially different when training on shuffled data. We evaluated all the models on the ordered DailyDialog test set to calculate perplexity. The minimal impact of shuffling the training data suggests that dialog models do not adequately leverage dialog structure during training. Our results show that essentially all of the information captured when training on ordered dialogs is also learned when training on shuffled dialogs.

\begin{table*}
\centering
\noindent\makebox[\textwidth]{
\begin{tabular}{l|cccccccc|c}\toprule
\multicolumn{1}{@{} l|}{Models} & TREC & DNLI & MWOZ & SGD & SNIPS & WNLI & SSA & Topic & Avg \\ \midrule

\multicolumn{1}{@{} l|}{\textbf{Scratch}} & -0.72&	-0.61&	-0.65&	-0.43	&-0.82&	-0.24&	-0.99&	0.40&	-0.75  \\
\multicolumn{1}{@{} l|}{\textbf{Pretrained}} &-0.76&	-0.80&	-0.74&	-0.81&	-0.71&	0.61&	-0.93&	0.65&	-0.76 \\
\multicolumn{1}{@{} l|}{\textbf{All}} & -0.55&	-0.84&	-0.71	&-0.87&	-0.63&	0.30&	-0.73	&-0.64&	-0.92 \\

\bottomrule
\end{tabular}}
\caption{Probing performance of the encoder state negatively correlates with test perplexity. Results imply that models with better data fit (lower perplexity) achieve better probing performance. Note that this is insufficient to establish a causal relationship.}
\label{tab:correlations}
\renewcommand{\arraystretch}{1.0}
\end{table*}

\section{Limitations}
Some of our conclusions assume that probing performance is indicative of performance on the end-task of dialog generation. Yet it could be the case that certain models learn high quality representations for probing but cannot effectively use them for generation, due to a weakness in the decoder for example. To address this limitation, future work could examine the relationship between probing performance and human judgements of conversation quality. \citet{belinkov2018internal} argues more research on the causal relation between probing and end-task performance is required to address this limitation. 

However, it is reasonable to assume that capturing information about a certain probing task is a pre-requisite to utilizing information relevant to that task for generation. For example, a model that cannot identify user sentiment is unlikely to use information about user sentiment for generation. We also find that lower perplexity (better data fit) is correlated with better probing performance (table \ref{tab:correlations}), suggesting that probing is a valuable, if imperfect, analysis tool for open-domain dialog systems.

\section{Conclusion}
We use probing to shed light on the conversational understanding of neural dialog models. Our findings suggest that standard neural dialog models suffer from many limitations. They do not effectively encode information about the conversation history, struggle to learn basic conversational skills, and fail to leverage the dyadic, turn-taking structure of dialog. These limitations are particularly severe for small models trained from scratch on dialog data but occasionally also affect large pre-trained models. Addressing these limitations is an interesting direction of future work. Models could be augmented with specific components or multi-task loss functions to support learning certain skills. Future work can also explore the relationship between probing performance and human evaluation.

\bibliography{acl2020}
\bibliographystyle{acl_natbib}

\newpage
\appendix

\section{Supplemental Material}
\label{sec:supplemental}

\subsection{Training Details}
For the small RNN trained from scratch, we used a 2-layer encoder, 2-layer decoder network with bidirectional LSTM units with a hidden size of 256 and a word embedding size of 128. For the small RNN with attention, we used the same architecture but also added multiplicative attention \cite{luong2015effective}. We set dropout to 0.3 and used a batch size of 64. We used an Adam optimizer \cite{kingma2014adam} with a learning rate of 0.005, inverse square root decay, and 4000 warm-up updates.

For the small Transformer, we used a 2-layer encoder, 2-layer decoder network with an embedding size of 400, 8 attention heads, and a feedforward network size of 300. We set dropout to 0.3 and used a batch size of 64.  We used an Adam optimizer with a learning rate of 0.001, inverse square root decay, and 6000 warm-up updates.

For the large RNN pretrained on Wikitext-103 \cite{merity2016wikitext}, we used a 2-layer encoder, 2-layer decoder network with bidirectional LSTM units with a hidden size of 1024 and a word embeddings size of 300. For the large RNN with attention, we used the same architecture but also included multiplicative attention. We set dropout to 0.3 and used a batch size of 40. We used an Adam optimizer with a learning rate of 0.005, inverse square root decay, and 4000 warm-up updates.

For the large Transformer we used a 2-layer encoder, 2-layer decoder network with an embedding size of 768, 12 attention heads, and a feedforward network size of 2048. We set dropout to 0.1 and used a batch size of 32.  We used an Adam optimizer with a learning rate of 0.001, inverse square root decay, and 4000 warm-up updates.

\subsection{Probing Tasks Examples}

Table \ref{tab:probingtasks} below, lists all the probing tasks and provides examples from each task. We also include the possible classes and training set sizes.

\begin{table*}
\centering
\noindent\makebox[\textwidth]{
\setcellgapes{0.2cm}
\makegapedcells
\begin{tabular}{p{1.8cm} c p{6.1cm} p{2.8cm} p{1.7cm}}
\toprule
\normalsize Dataset  & $|$Train$|$ & Example & Classes & Label  \\ 
\midrule

TREC & 5.5K &
\small
\makecell[lp{6cm}]{{[}Usr1{]}: Why do heavier objects travel downhill faster?}
& \makecell[cl]{entity, number\\ description,\\location,  \dots}
& description
\\\hline

Dialogue NLI & 310K &
\small
\makecell[lp{6cm}]{{[}Usr1{]}: I go to college part time. \\
{[}Usr2{]}: You are a recent college graduate looking for a job.}
& \makecell[cl]{entail,\\ contradict,\\neutral}
& contradict
\\\hline

MultiWOZ & 8.5K &
\small
\makecell[lp{6cm}]{{[}Usr1{]}: I need to book a hotel.\\ {[}Usr2{]}: I can help you with that. What is your price range? \\
{[}Usr1{]}: That doesn't matter as long as it has free wifi and parking.}
& \makecell[cl]{hotel-inform, \\taxi-request, \\general-thank, \\\dots}
& hotel-inform
\\\hline

Schema- Guided & 16K &
\small
\makecell[lp{6cm}]{{[}Usr1{]}: Help me find a restaurant.\\ 
{[}Usr2{]}: Which city are you looking in?\\
{[}Usr1{]}: Cupertino, please. }
& \makecell[cl]{find-restaurant,\\ get-ride, \\ reserve-flight, \\ \dots}
& find-restaurant
\\\hline

SNIPS & 14K &
\small
\makecell[lp{6cm}]{{[}Usr1{]}: I want to see Outcast.}
& \makecell[cl]{search-screening,\\ play-music, \\ get-weather, \\ \dots}
& search-screening
\\\hline

Winograd NLI & 0.6K &
\small
\makecell[lp{6cm}]{[User1]: John couldn't see the stage with Billy in front of him because he is so tall.
\\ {[}User2{]}: John is so tall.}
& \makecell[cl]{entail,\\ contradict}
& contradict
\\\hline

ScenrioSA & 1.9K &
\small
\makecell[lp{6cm}]{{[}Usr1{]}: Thank you for coming, officer. \\
{[}Usr2{]}: What seems to be the problem? \\
{[}Usr1{]}: I was in school all day and came home to a burglarized apartment.}
& \makecell[cl]{positive, \\ negative, \\ neutral }
& negative
\\\hline

DailyDialog Topic & 0.9K &
\small
\makecell[lp{6cm}]{{[}Usr1{]}: I think Yoga is suitable for me.\\ 
{[}Usr2{]}: Why?\\
{[}Usr1{]}: Because it doesn't require a lot of energy.\\
{[}Usr2{]}: But I see people sweat a lot doing Yoga too. }
& \makecell[cl]{ordinary life, \\ work, school, \\ tourism, politics, \\relationship, ...}
& ordinary life
\\\bottomrule

\end{tabular}}
\caption{Examples from the selected probing tasks.}
\label{tab:probingtasks}
\end{table*}

\end{document}